\documentclass[journal=jacsat,manuscript=article]{achemso}

\usepackage[version=3]{mhchem} 
\usepackage{multirow}
\usepackage{makecell}
\usepackage{cellspace}
\usepackage{booktabs}
\usepackage{amsmath} 
\usepackage{amssymb} 
\usepackage{enumitem}
\usepackage{algorithm}
\usepackage{algpseudocode}

\author{Janghoon Ock}
\affiliation{Department of Chemical Engineering, Carnegie Mellon University, 5000 Forbes Ave, Pittsburgh, PA 15213, USA}
\author{Radheesh Sharma Meda}
\affiliation{Department of Mechanical Engineering, Carnegie Mellon University, 5000 Forbes Ave, Pittsburgh, PA 15213, USA}
\author{Tirtha Vinchurkar}
\affiliation{Department of Chemical Engineering, Carnegie Mellon University, 5000 Forbes Ave, Pittsburgh, PA 15213, USA}
\author{Yayati Jadhav}
\affiliation{Department of Mechanical Engineering, Carnegie Mellon University, 5000 Forbes Ave, Pittsburgh, PA 15213, USA}
\author{Amir Barati Farimani}
\affiliation{Department of Mechanical Engineering, Carnegie Mellon University, 5000 Forbes Ave, Pittsburgh, PA 15213, USA}
\email{barati@cmu.edu}

\title[An \textsf{achemso} demo]
  {Adsorb-Agent: Autonomous Identification of Stable Adsorption Configurations via Large Language Model Agent}

\keywords{Large Language Model, AI Agent, Computational Catalysis, AI Agent \LaTeX}

\begin{document}



\begin{abstract}
    Adsorption energy is a key reactivity descriptor in catalysis. Determining adsorption energy requires evaluating numerous adsorbate-catalyst configurations, making it computationally intensive. Current methods rely on exhaustive sampling, which does not guarantee the identification of the global minimum energy. To address this, we introduce Adsorb-Agent, a Large Language Model (LLM) agent designed to efficiently identify stable adsorption configurations corresponding to the global minimum energy. Adsorb-Agent leverages its built-in knowledge and reasoning to strategically explore configurations, significantly reducing the number of initial setups required while improving energy prediction accuracy. In this study, we also evaluated the performance of different LLMs—GPT-4o, GPT-4o-mini, Claude-3.7-Sonnet, and Deepseek-Chat—as the reasoning engine for Adsorb-Agent, with GPT-4o showing the strongest overall performance. Tested on twenty diverse systems, Adsorb-Agent identifies comparable adsorption energies for 84\% of cases and achieves lower energies for 35\%, particularly excelling in complex systems. It identifies lower energies in 47\% of intermetallic systems and 67\% of systems with large adsorbates. These findings demonstrate Adsorb-Agent’s potential to accelerate catalyst discovery by reducing computational costs and enhancing prediction reliability compared to exhaustive search methods.
\end{abstract}

\section{Introduction}

The design of optimal catalyst materials for targeted reaction processes plays an essential role in advancing chemical processes~\cite{Norskov2014, Michel2008, oc20_intro}. In particular, addressing the dual challenge of meeting the growing global energy demand while combating climate change necessitates the development of efficient, low-cost catalysts to enable the broader use of renewable energy sources~\cite{oc20_intro}. Traditionally, the search for optimal catalysts has relied on either labor-intensive experimental methods or computationally expensive quantum chemistry calculations. Because of the vast material design space, screening strategies often focus on identifying key descriptors that effectively predict catalytic performance.

Adsorption energy, defined as the change in energy upon the binding of a molecule to a catalytic surface, is one of the most widely used descriptors in computational catalysis due to its direct correlation with catalytic reactivity~\cite{Norskov2009, Yang2014}. The adsorption energy, corresponding to the most stable adsorption configuration, serves as a key descriptor of catalyst performance and plays a crucial role in estimating the reactivity of various catalysts~\cite{Norskov2009, bep, bep2, Ulissi2017}. Furthermore, adsorption energy is a foundational parameter in constructing free energy diagrams, which are used to identify the energetically preferred reaction pathways on catalyst surfaces.

The adsorption energy, $\Delta E_{\text{ads}}$, is mathematically defined as the global minimum energy among all possible adsorption configurations~\cite{ock2023error, adsorbml}. It is calculated as the difference between the total energy of the adsorbate-catalyst system ($E_{\text{sys}, i}$), the energy of the clean surface (slab) ($E_{\text{slab}}$), and the energy of the gas-phase adsorbate or reference species ($E_{\text{gas}}$):

\begin{subequations}
\label{eq:Eads}
\begin{align}
\Delta E_i &= E_{\text{sys}, i} - E_{\text{slab}} - E_{\text{gas}} \label{eq:Ei} \\
\Delta E_{\text{ads}} &= \min_i (\Delta E_i) \label{eq:Eads-min}
\end{align}
\end{subequations}

Accurately determining adsorption energy presents significant challenges. The complex electron-level interactions that govern chemical bonding make it impractical to predict the most stable configuration based solely on atomic-level information. As a result, determining the global minimum adsorption energy typically requires enumerating and evaluating a vast number of possible configurations~\cite{adsorbml, catkit, pymatgen}. This process becomes computationally prohibitive when using quantum chemistry methods such as density functional theory (DFT)~\cite{adsorbml, adsorbdiff}. The difficulty is further compounded by the combinatorial explosion of potential binding sites, variations in surface geometries, and diverse orientations the adsorbate can adopt. Despite exhaustive configuration searches, there is no guarantee of reliably identifying the true global minimum energy configuration. These challenges underscore the need for more efficient and accurate approaches to streamline adsorption energy determination, paving the way for faster and more reliable catalyst design.

Recent advances in machine learning (ML) have introduced promising alternatives to conventional quantum chemistry methods, significantly improving the efficiency of adsorption energy prediction tasks. In particular, Graph Neural Networks (GNNs) have demonstrated exceptional performance in predicting energy and forces for atomic systems. For adsorbate-catalyst systems~\cite{oc20}, GNNs achieve a high level of precision, predicting adsorption energies with an error of approximately 0.2 eV and forces with an error of 0.013 eV/Å~\cite{ocp_lb, equiformerv2}. These capabilities make GNNs effective surrogates for DFT calculations in tasks such as geometry optimization and energy prediction. Building on this foundation, Lan et al. introduced the AdsorbML method, which achieved a 2000$\times$ speedup in adsorption energy determination while retaining 87.36\% of the accuracy of full DFT calculations by integrating GNNs with DFT~\cite{adsorbml}. In their approach, GNNs are used to relax structures from initial adsorption configurations, after which DFT is employed for further relaxation or single-point energy calculations to obtain DFT-validated adsorption energies. Despite this progress, the placement of adsorbates on the surface and the sampling of adsorption sites remain reliant on exhaustive enumeration, posing a significant challenge in reducing the initial search space and improving computational efficiency. 


Moving forward, large language models (LLMs) are emerging as transformative tools in scientific problem-solving by leveraging their pre-trained knowledge and planning capabilities, as well as their seamless human-machine interaction~\cite{aiscientist}. LLM agents, powered by LLMs, have demonstrated remarkable potential in reshaping scientific workflows. For example, Boiko et al. proposed Coscientist, an LLM agent that automates experimental design and execution, significantly enhancing productivity while reducing manual effort~\cite{coscientist}. Similarly, Szymanski et al. demonstrated the application of LLMs in an autonomous laboratory framework for proposing synthesis recipes~\cite{alab}. In the field of additive manufacturing, Jadhav et al. introduced the LLM-3D Print framework, which streamlines the design-to-manufacturing process by autonomously generating, validating, and optimizing 3D printing instructions with an LLM agent at its core~\cite{llm3dprint}. These breakthroughs demonstrate how LLMs are reshaping the landscape of scientific discovery and innovation.

In this study, we introduce the Adsorb-Agent, an LLM-based agent designed to efficiently determine adsorption energy. The Adsorb-Agent predicts initial adsorption configurations that are likely to be close to the most stable configuration and relaxes them to identify the minimum energy state. While human researchers can propose plausible stable adsorption configurations for specific systems based on domain knowledge—such as chemical bonding and surface science—it remains exceedingly difficult to derive a universal theorem for predicting the most stable configuration across diverse adsorbate-catalyst systems. Moreover, in high-throughput screening, where millions of candidate systems must be evaluated, it is infeasible to manually propose stable configurations for each individual system~\cite{catlas}. The Adsorb-Agent addresses these challenges by autonomously deriving stable configurations, relying solely on the LLM’s built-in knowledge and emergent reasoning capabilities. Because it operates purely through inference from pre-trained models, it is readily applicable to large-scale tasks.



This study has two primary objectives: first, to reduce the computational cost of adsorption energy identification by minimizing the number of initial configurations required; and second, to enhance the accuracy of adsorption energy predictions by generating refined initial configurations that are closer to the global minimum, while maintaining human interpretability. By bridging state-of-the-art LLM capabilities with catalytic configuration challenges, the Adsorb-Agent represents a significant step toward broader adoption of AI-driven methods in materials science and catalysis, accelerating the discovery and design of optimal catalysts. Furthermore, when integrated with other LLM models and tools for catalyst design~\cite{ock2023catberta, ock2024multimodal,adsorbdiff}, the Adsorb-Agent can be extended to a wider range of applications in optimal catalyst development.

\section{Agentic Framework}
\subsection{Workflow Overview}
\begin{figure*}[!htb] 
\centering
\includegraphics[width=0.99\textwidth]{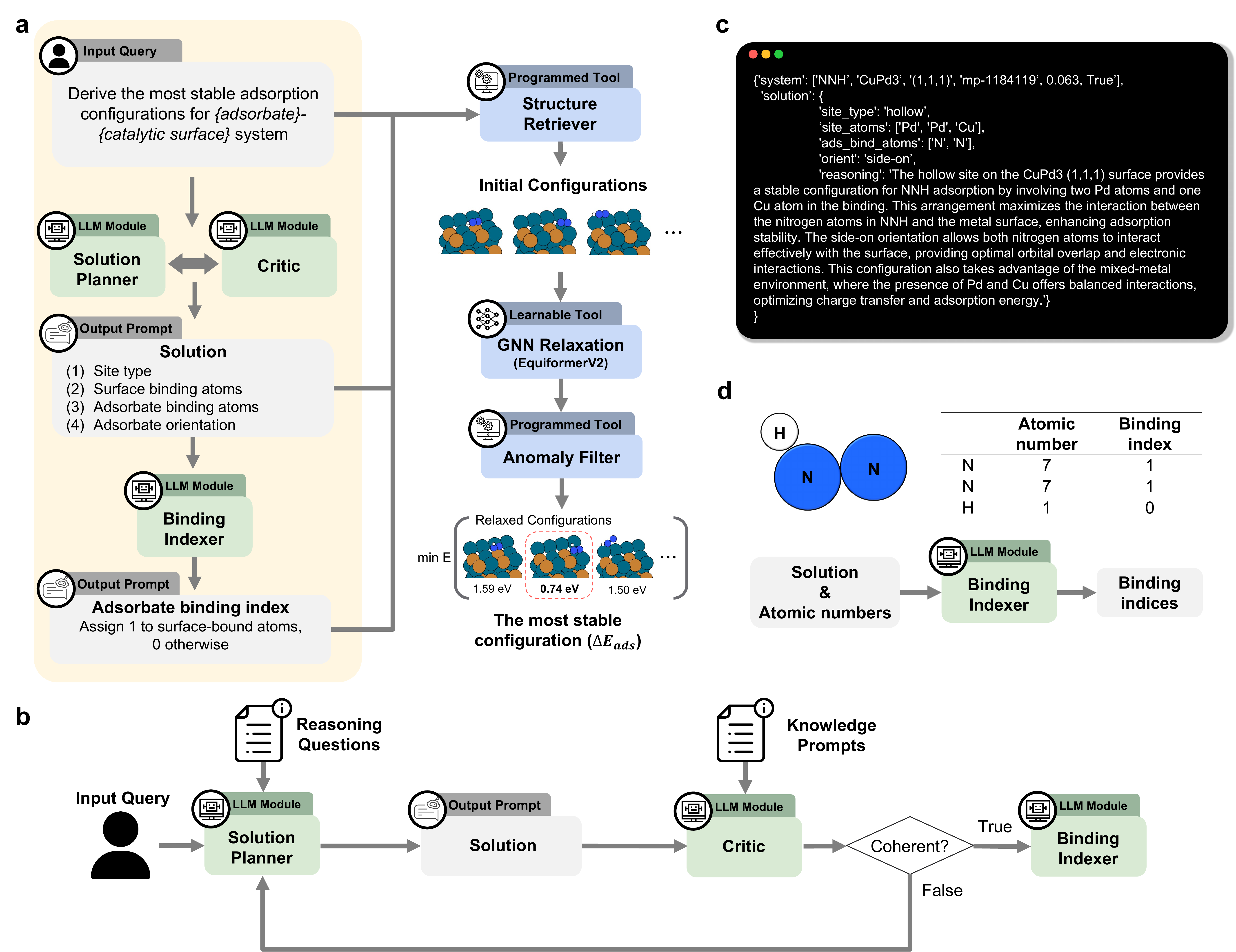} 
\caption{Adsorb-Agent Framework.  
\textbf{a.} Overall process for identifying the most stable configuration by combining the LLM modules and tools. The left yellow box represents the part where the LLM modules are involved, comprising three components—Solution Planner, Critic, and Binding Indexer. These modules are powered by LLM APIs such as GPT-4o, GPT-4o-mini, Claude-3.7-Sonnet, and Deepseek-Chat. \textbf{b.} Illustration of the iterative interaction between the Solution Planner and Critic modules. \textbf{c.} Example solution for the \ce{NNH}-\ce{CuPd3} (1,1,1) system. The system metadata comprises the following, in order: the SMILES representation of the adsorbate molecule, the chemical composition of the bulk catalyst, the Miller index, the Materials Project ID, the shift, and the top.  \textbf{d.} Example of binding index derivation from the Binding Indexer module.}
\label{fig:framework}
\end{figure*}

The Adsorb-Agent is an LLM-powered agent designed to identify the most stable adsorption configuration and its corresponding adsorption energy. It consists of three core LLM modules—the Solution Planner, Critic, and Binding Indexer—all powered by LLMs, as illustrated in Fig.~\ref{fig:framework}a. The process begins with a user query that specifies the adsorbate’s SMILES, the catalyst’s chemical symbol, and the surface orientation. The core functionality of Adsorb-Agent is to narrow the configuration search space by identifying promising adsorption candidates for the specified adsorbate-catalyst system. The agent retrieves the adsorbate molecule and catalyst surface structure from a database, places the adsorbate according to the predicted configuration, and then relaxes the system using designated computational tools to calculate the adsorption energy. By comparing the relaxed energies, the agent identifies the configuration with the lowest energy, which serves as a key reactivity descriptor in catalysis.

In this work, LangChain is used to manage interactions between the LLM and the various modules of the Adsorb-Agent framework. LangChain enables efficient handling of complex workflows involving LLMs by organizing prompts, processing responses, and integrating external tools or APIs~\cite{chase2022langchain}. 

\subsection{Configuration Candidates}

Based on the natural language input query, the Solution Planner derives the description for the most probable stable adsorption configuration. This process is primarily conducted based on the LLM's built-in knowledge and reasoning capability. The solution output includes four pieces of information: the type of adsorption site, the binding atoms on both the surface and the adsorbate, and the orientation of the adsorbate. This derivation is guided by a structured list of reasoning questions provided by the user, as shown in Fig.~\ref{fig:framework}b. These questions reflect the typical thought process of human researchers—for example, ``Are the bonds between the adsorbate and surface strong enough to ensure stability?"—while deliberately excluding system-specific knowledge to ensure general applicability (see Supplementary Fig. S1). The Solution Planner can also provide reasoning statements that explain how it arrived at its prediction, as illustrated in Figure 1c. A full set of outputs for 20 systems is included in the Supplementary Information.

To ensure the logical coherence of the solution output, the Critic module evaluates the initial solution generated by the Solution Planner. It takes the solution as input and uses a knowledge prompt that clarifies the terms within the solution to guide its review (see Fig.~\ref{fig:framework}b). Details of the knowledge prompt are provided in Supplementary Fig. S1. The module focuses on two aspects: (1) the coherence between the adsorption site type and the binding atoms on the surface, and (2) the alignment between the adsorbate’s binding atoms and its orientation. For instance, if the solution specifies a bridge site, it must involve two binding surface atoms. Similarly, for adsorbate orientation, if the adsorbate is described as end-on, it should have only one binding adsorbate atom. If any incoherence is identified, the Critic rebuts the solution, and the Solution Planner is re-initialized to produce a revised solution. This iterative interaction ensures the final adsorption configuration is logically coherent and self-consistent.

Once a coherent solution is generated, the Binding Indexer module assigns indices to the adsorbate’s binding atoms based on the solution. This step translates the human-readable configuration into a numerical format suitable for computational processing. As illustrated in Fig.~\ref{fig:framework}d, the Binding Indexer takes the identified binding atoms and orientation from the solution, along with the adsorbate's atomic number array, to generate a binding index array. This array specifies which atoms in the adsorbate are involved in binding to the surface. Using this array, the adsorbate can be positioned on the catalytic surface, reflecting both the surface orientation and binding atom information. This automation removes the need for manually pre-defining binding atoms—commonly required in datasets like the Open Catalyst Project, which explicitly mark binding atoms with asterisks (e.g., NNH). In addition, we introduce a new placement strategy capable of handling side-on adsorbates (see Methods section).

\subsection{Energy Determination}
The following steps are carried out without the involvement of the LLM modules, but using the pre-written computation scripts. The catalytic surface and adsorbate molecule are retrieved using the Open Catalyst Project demo API, based on the adsorbate SMILES, catalyst bulk composition, and Miller indices provided in the user query. Details about the Open Catalyst Project demo are provided in the Methods section. The adsorbate is then placed onto the catalytic surface to generate initial adsorbate–catalyst structures, guided by the predicted configuration and the binding index array. As illustrated in the atomic visualization in Fig.~\ref{fig:framework}, multiple initial configurations remain possible; however, their number is significantly reduced compared to conventional configuration enumeration approaches. 

These initial structures are subsequently relaxed to determine the minimum energy configuration, as relaxed energies are necessary for meaningful comparison. In this study, we employ a GNN model, specifically EquiformerV2 trained on the Open Catalyst 2020 dataset~\cite{oc20, equiformerv2}, although other machine learning models or qunantum chemistry simulations could also be used. Details of the GNN-based relaxation process are provided in the Methods section. During the relaxation process, anomalies such as extensive surface reconstruction, adsorbate dissociation, or desorption may occur, and any structures exhibiting these anomalies are filtered out~\cite{adsorbml, peterson2014global, jung2023machine}. Among the remaining valid configurations, the one with the lowest energy is identified as the most stable configuration. This energy is recognized as the adsorption energy, which serves as a critical reactivity descriptor for the adsorbate-catalyst combination.

\subsection{Performance Evaluation}

Theoretically, the most stable adsorption configuration corresponds to the global minimum energy among all possible configurations. However, due to the vast configurational space, it is practically impossible to exhaustively identify the true global minimum. Conventional approaches instead rely on enumeration algorithms, such as \texttt{random} and \texttt{heuristic} algorithms, which sample numerous configurations to approximate the global minimum adsorption energy. Details of these enumeration procedures are provided in the Methods section.

The Adsorb-Agent’s ability to identify the most stable configuration is evaluated against conventional enumeration strategies. Performance is assessed based on three key criteria: (i) efficiency in reducing the configuration search space, (ii) accuracy in identifying adsorption configurations with energies comparable to those found by enumeration algorithms, and (iii) consistency of results across independent trials.

The evaluation is conducted on 20 adsorbate–catalyst systems selected for their practical importance, particularly in nitrogen production and fuel cell applications~\cite{NRR_1, NRR_2, ORR_1}. For example, the electrocatalytic nitrogen reduction reaction (NRR) offers a potential route for sustainable nitrogen fixation. However, its performance is often limited by the high activation energy required to cleave the inert \ce{N#N} bond and by strong competition from the hydrogen evolution reaction (HER)\cite{zhou2023enhanced}. Likewise, the oxygen reduction reaction (ORR) is a vital reaction process in fuel cell operation and holds a central position in the broader field of electrocatalysis~\cite{ORR_1}.

Eight of the selected systems are associated with NRR and HER. These include four catalysts proposed by Zhou et al.\cite{zhou2023enhanced}, each interacting with two key adsorbates: \ce{NNH} for NRR and \ce{H} for HER, resulting in a total of eight systems. An additional six systems are associated with ORR, with \ce{OH} as the key adsorbate, selected from the experimentally verified sets reported by Kulkarni et al.\cite{ORR_1, ORR_2, ORR_3}. To broaden the evaluation beyond simple adsorbates, six more systems featuring larger molecules—such as \ce{CH2CH2OH}, \ce{OCHCH3}, and \ce{ONN(CH3)2}—and intermetallic catalysts were included. These complex systems were randomly selected from the Open Catalyst 2020-Dense (OC20-Dense) dataset~\cite{adsorbml}. A complete list of the adsorbate–catalyst systems is provided in Table~\ref{tab:results}, with detailed slab information available in Supplementary Table S1.






\section{Results and discussion}
\subsection{Search Space Reduction Demonstration}
\begin{figure*}[!htb]
\centering
\includegraphics[width=0.9\textwidth]{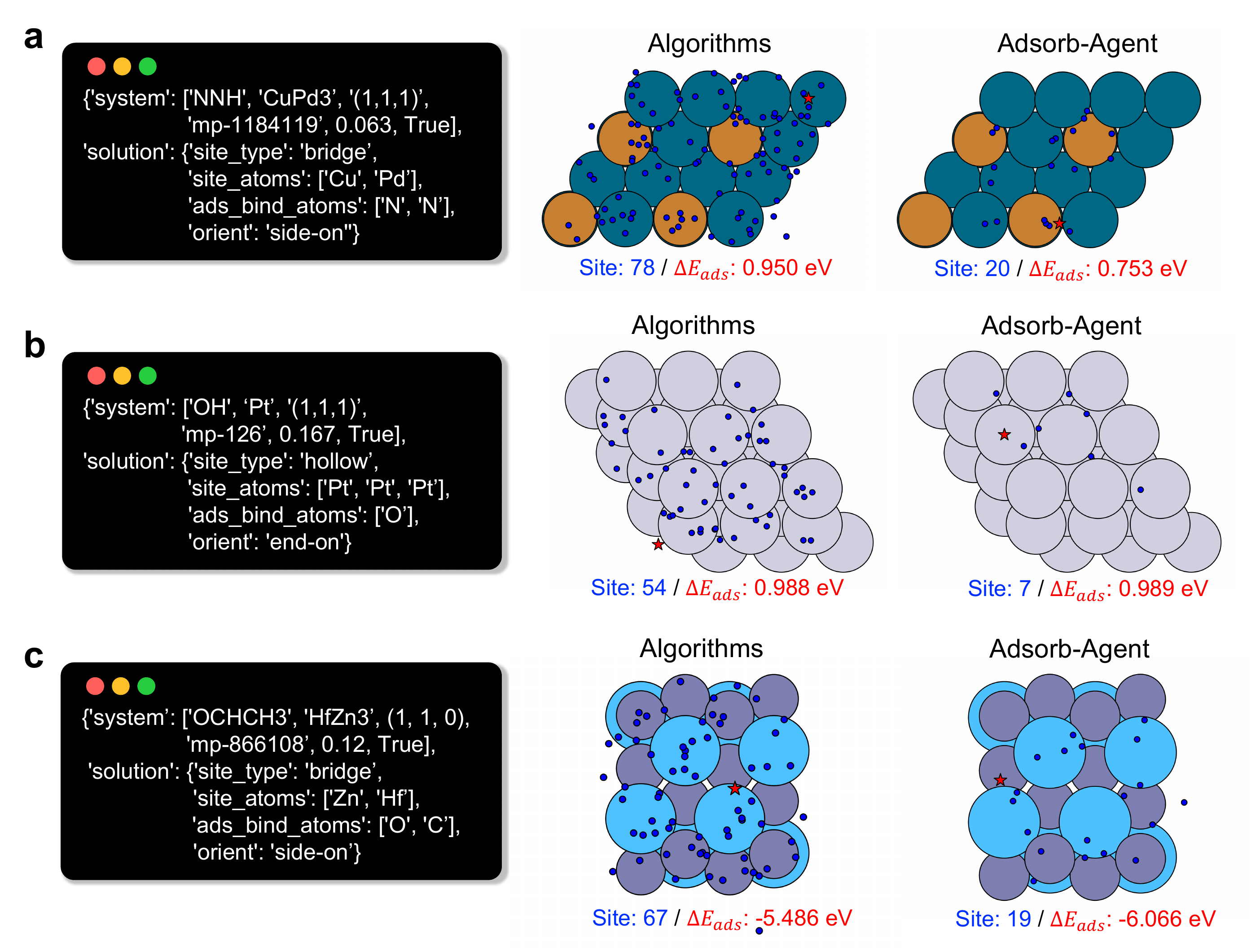}
\caption{Comparison of Search Space Between the Algorithmic Approach and the Adsorb-Agent. Left panels show solutions derived by the Adsorb-Agent, while right panels present atomic visualizations of surfaces with initial adsorption sites for \textbf{a.} \ce{NNH}-\ce{CuPd3} (111), \textbf{b.} \ce{OH}-\ce{Pt} (111), and \textbf{c.} \ce{OCHCH3}-\ce{HfZn3} (110). (Blue dots: initial sites; red stars: relaxed stable sites)}
\label{fig:inits}
\end{figure*}

The Adsorb-Agent effectively reduces the search space for further energy determination by specifying both the adsorption site and the binding atoms, thereby limiting the number of initial configurations to evaluate. In search problems, beginning from an initial point proximal to the optimal solution is essential, as this can significantly reduce the search space and improve convergence efficiency\cite{ock2024gradnav}. Fig.~\ref{fig:inits} illustrates three example cases: \ce{NNH}-\ce{CuPd3} (111), \ce{OH}-\ce{Pt} (111), and \ce{OCHCH3}-\ce{HfZn3} (110), which represent NRR-related, ORR-related, and large adsorbate-containing systems, respectively. Among these, the \ce{OCHCH3}-\ce{HfZn3} system serves as an example of a monometallic surface, which is more homogeneous compared to intermetallic surfaces. In the atomic visualizations, blue dots denote the initial adsorption sites, while red stars mark the relaxed adsorption sites corresponding to the most stable configurations.

The Adsorb-Agent significantly reduces the number of initial configurations by focusing on likely adsorption sites, compared to the exhaustive enumeration algorithmic approaches. For example, the Adsorb-Agent predicts the bridge site at the intersection of \ce{Cu} and \ce{Pd} atoms as the primary adsorption site for \ce{NNH}-\ce{CuPd3} (111). Similarly, it identifies the hollow site of the \ce{Pt} (111) surface as the optimal adsorption site for \ce{OH} and the bridge site between \ce{Zn} and \ce{Hf} atoms for the \ce{OCHCH3}-\ce{HfZn3} (110) system. Occasional deviations from the predicted solutions arise due to the distance margin used to define targeted adsorption sites. By focusing on specific sites, the Adsorb-Agent effectively reduces the search space.

For the \ce{NNH}-\ce{CuPd3} (111) and \ce{OCHCH3}-\ce{HfZn3} (110) systems, the relaxed adsorption sites of the most stable configurations precisely match the solutions proposed by the Adsorb-Agent. In both cases, the Adsorb-Agent achieves adsorption energies lower than those obtained via algorithmic approaches. However, for the \ce{OH}-\ce{Pt} (111) system, the adsorption site of the most stable relaxed configuration identified by both the algorithmic methods and the Adsorb-Agent is an ontop site, differing from the initial Adsorb-Agent prediction. Despite this discrepancy, the initial site suggested by the Adsorb-Agent successfully guided the system to the most stable configuration during the relaxation process.


\subsection{Adsorption Energy Identification}
The Adsorb-Agent efficiently identifies the most stable configurations while significantly reducing the size of the search space. Its performance was evaluated against conventional algorithmic approaches (\texttt{random} and \texttt{heuristic}), as summarized in Table~\ref{tab:results}.  

\begin{table}[!htbp]
\centering
\caption{Comparison of Adsorb-Agent and Algorithmic Approach. The adsorption energy corresponds to the minimum energy among configurations. Results for individual runs are provided in Supplementary Table S1.}
\label{tab:results}
\resizebox{\textwidth}{!}{%
\begin{tabular}{ccccccc}
\toprule
\textbf{No.} & \textbf{Adsorbate} & \textbf{Catalyst} & \multicolumn{2}{c}{\textbf{Adsorption Energy [eV]}} & \multicolumn{2}{c}{\textbf{Number of Initial Sets}} \\ 
\cmidrule(lr){4-5} \cmidrule(lr){6-7} &  &  & \textbf{Adsorb-Agent (\(\downarrow\))} & \textbf{Algorithm} & \textbf{Adsorb-Agent (\(\downarrow\))} & \textbf{Algorithm} \\ 
\midrule
1   & H   & \ce{Mo3Pd} (111) & -0.764 \(\pm\) 0.113 & -0.941 \(\pm\) 0.002 & 6.7 \(\pm\) 2.1  & 59 \\
2   & NNH & \ce{Mo3Pd} (111) & -1.265 \(\pm\) 0.158 & -0.903 \(\pm\) 0.117 & 9.3 \(\pm\) 3.7  & 51 \\ 
3   & H   & \ce{CuPd3} (111) & -0.380 \(\pm\) 0.003 & -0.398 \(\pm\) 0.017 & 16.7 \(\pm\) 1.2 & 98 \\ 
4   & NNH & \ce{CuPd3} (111) &  0.745 \(\pm\) 0.006 &  0.867 \(\pm\) 0.072 & 17.3 \(\pm\) 3.1 & 78 \\ 
5   & H   & \ce{Cu3Ag} (111) &  -0.019 \(\pm\) 0.041 & -0.072 \(\pm\) 0.002 & 21.3 \(\pm\) 4.1& 98 \\ 
6   & NNH & \ce{Cu3Ag} (111) &  1.504 \(\pm\) 0.057 &  1.500 \(\pm\) 0.002 & 16.7 \(\pm\) 2.6 & 56 \\ 
7   & H   & \ce{Ru3Mo} (111) & -0.587 \(\pm\) 0.002 & -0.586 \(\pm\) 0.050 & 17.0 \(\pm\) 2.2 & 94 \\ 
8   & NNH & \ce{Ru3Mo} (111) & -0.498 \(\pm\) 0.013 & -0.276 \(\pm\) 0.003 & 18.7 \(\pm\) 0.5 & 81 \\ 
9   & OH  & \ce{Pt} (111)    &  0.990 \(\pm\) 0.001 &  0.990 \(\pm\) 0.071 & 7.0 \(\pm\) 1.6  & 54 \\ 
10  & OH  & \ce{Pt} (100)    &  0.991 \(\pm\) 0.001 &  0.991 \(\pm\) 0.001 & 10.3 \(\pm\) 4.2 & 54 \\ 
11  & OH  & \ce{Pd} (111)    &  0.814 \(\pm\) 0.000 &  0.814 \(\pm\) 0.001 & 20.0 \(\pm\) 5.7 & 54 \\ 
12  & OH  & \ce{Au} (111)    &  1.408 \(\pm\) 0.002 &  1.409 \(\pm\) 0.002 & 23.3 \(\pm\) 5.2 & 54 \\ 
13  & OH  & \ce{Ag} (100)    &  0.440 \(\pm\) 0.001 &  0.463 \(\pm\) 0.009 & 23.7 \(\pm\) 4.5 & 53 \\ 
14  & OH  & \ce{CoPt} (111)  & -0.208 \(\pm\) 0.015 & -0.166 \(\pm\) 0.046 & 41.3 \(\pm\) 1.2 & 120\\ 
15  & \ce{CH2CH2OH} & \ce{Cu6Ga2} (100) & -2.338 \(\pm\) 0.833 & -3.077 \(\pm\) 0.062 & 28.7 \(\pm\) 15.5& 66 \\ 
16  & \ce{CH2CH2OH} & \ce{Au2Hf} (102)  & -2.761 \(\pm\) 0.592 & -3.761 \(\pm\) 0.129 & 28.0 \(\pm\) 4.5 & 78 \\ 
17  & \ce{OCHCH3}   & \ce{Rh2Ti2} (111) & -4.561 \(\pm\) 0.007 & -4.275 \(\pm\) 0.086 & 29.0 \(\pm\) 4.3 & 62 \\ 
18  & \ce{OCHCH3}   & \ce{Al3Zr} (101)  & -4.616 \(\pm\) 0.014 & -4.325 \(\pm\) 0.052 & 22.0 \(\pm\) 2.4 & 68 \\ 
19  & \ce{OCHCH3}   & \ce{Hf2Zn6} (110) & -5.922 \(\pm\) 0.209 & -5.443 \(\pm\) 0.037 & 18.0 \(\pm\) 2.2 & 67 \\ 
20  & \ce{ONN(CH3)2}& \ce{Bi2Ti6} (211) & -3.454 \(\pm\) 0.337 & -2.441 \(\pm\) 0.103 & 33.0 \(\pm\) 3.6 & 139\\ 
\bottomrule
\end{tabular}
}
\end{table}

To quantify the effectiveness of the Adsorb-Agent in identifying adsorption energies, three key metrics are defined, with their mathematical formulations provided in the Methods section. Success Ratio (SR) assesses the ability of the Adsorb-Agent to identify adsorption energies comparable to those found by the algorithmic approaches. Lower Energy Discovery Ratio (LEDR) measures the capability of the Adsorb-Agent to discover adsorption energies lower than those identified by the algorithmic approaches. Reduced Search Space Ratio (RSR) quantifies the reduction in the number of initial configurations required by the Adsorb-Agent compared to the algorithmic approaches. 

These metrics provide a comprehensive framework for evaluating the Adsorb-Agent's effectiveness in identifying the most stable configurations relative to conventional methods. Specifically, RSR reflects the efficiency of the search process, while SR and LEDR reflect the accuracy of energy identification. Notably, an increase in the number of initial configurations, as reflected by a higher RSR, often corresponds to improvements in SR and LEDR values. Therefore, all three metrics should be considered collectively to thoroughly assess the Adsorb-Agent’s performance in comparison to algorithmic approaches.

\begin{figure*}[!htb] 
\centering
\includegraphics[width=0.8\textwidth]{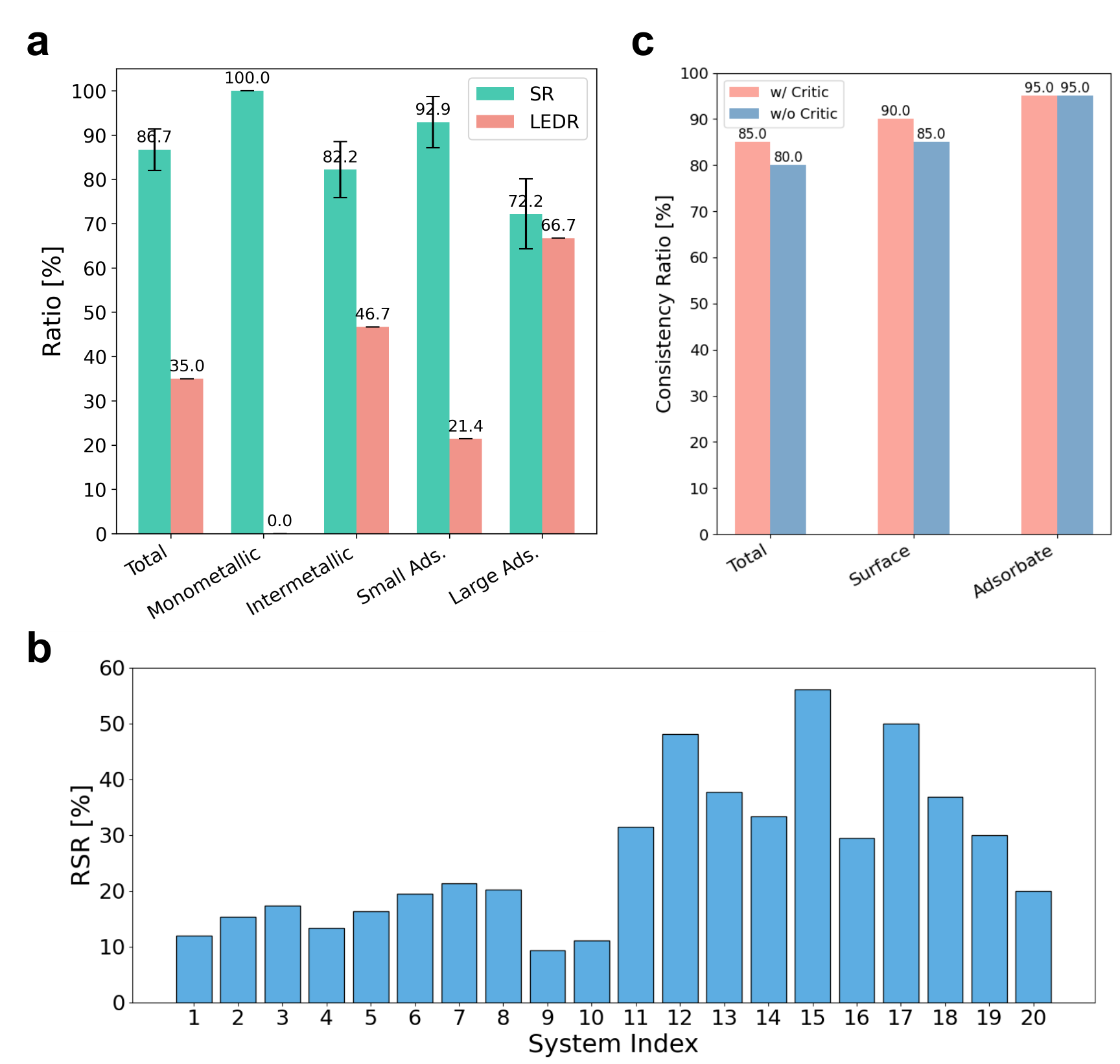} 
\caption{Quantitative Analysis of Adsorb-Agent Results. \textbf{a.} Success Rate (SR) and Lower Energy Discovery Rate (LEDR) across different categories; \textbf{b.} Reduced Search Space Rate (RSR) for the first run as an example; \textbf{c.} Consistency Rate across independent runs.}
\label{fig:metric}
\end{figure*}

As shown in Fig.~\ref{fig:metric}a, the Adsorb-Agent successfully identifies adsorption energies comparable to those found by the algorithmic approach in 83.7\% of cases and discovers lower energies in 35.0\% of cases. Remarkably, it achieved these results while using only 6.8–63.6\% of the initial configurations required by the algorithmic methods (see Fig.~\ref{fig:metric}b). As discussed earlier, increasing the number of initial configurations is likely to improve both the SR and LEDR. To ensure a fair comparison in this study, the number of initial configurations used by the Adsorb-Agent was scaled relative to the algorithmic approach, resulting in a reduction to 6.8–63.6\% of the original, with an average of 26.9\%. For practical applications, the number of initial configurations can be further increased, particularly for systems with lower RSRs, to improve the performance of adsorption energy determination.

An analysis of specific system categories reveals that the results vary with system complexity, as shown in Fig.~\ref{fig:metric}a. The systems are categorized based on the composition of the catalytic surface (monometallic and intermetallic) and the size of the adsorbate molecule (small and large molecules). Large adsorbates are defined as those with more than three atoms, such as \ce{CH2CH2OH}, \ce{OCHCH3}, and \ce{ONN(CH3)2}.

For systems with monometallic catalysts, the Adsorb-Agent consistently identifies adsorption energies comparable to those found by the algorithmic approach across all three trials. However, it does not achieve lower adsorption energies, indicating that the algorithmic approach successfully identifies the adsorption configuration with the global minimum energy. This outcome is likely due to the relatively homogeneous nature of monometallic surfaces, which consist of a single atom type (see Fig.~\ref{fig:inits}b). These findings suggest that extensive site enumeration is unnecessary for monotonous surfaces.

In contrast, for systems with intermetallic catalysts, the Adsorb-Agent demonstrates a distinct advantage. While the SR slightly decreases to 82.2\% compared to monometallic systems, the LEDR significantly improves to 46.7\%. This highlights the ability of the Adsorb-Agent to uncover new global minima through targeted searches, which the algorithmic approach cannot achieve through simple enumeration because of the increased complexity and heterogeneity of intermetallic surfaces.

A similar trend is observed when analyzing systems based on adsorbate complexity. Systems with large adsorbates exhibit a lower SR but a significantly higher LEDR compared to those with small adsorbates. Notably, systems with large adsorbates achieve the highest LEDR of 66.7\%, underscoring the effectiveness of the targeted search approach employed by the Adsorb-Agent. These results suggest that simple enumeration is relatively less effective at identifying the global minimum in systems containing complex adsorbates. Furthermore, this finding reinforces the effectiveness of the Adsorb-Agent in addressing these challenges.

\subsection{Consistency Across Independent Trials}
\begin{figure*}[!hptb] 
\centering
\includegraphics[width=0.99\textwidth]{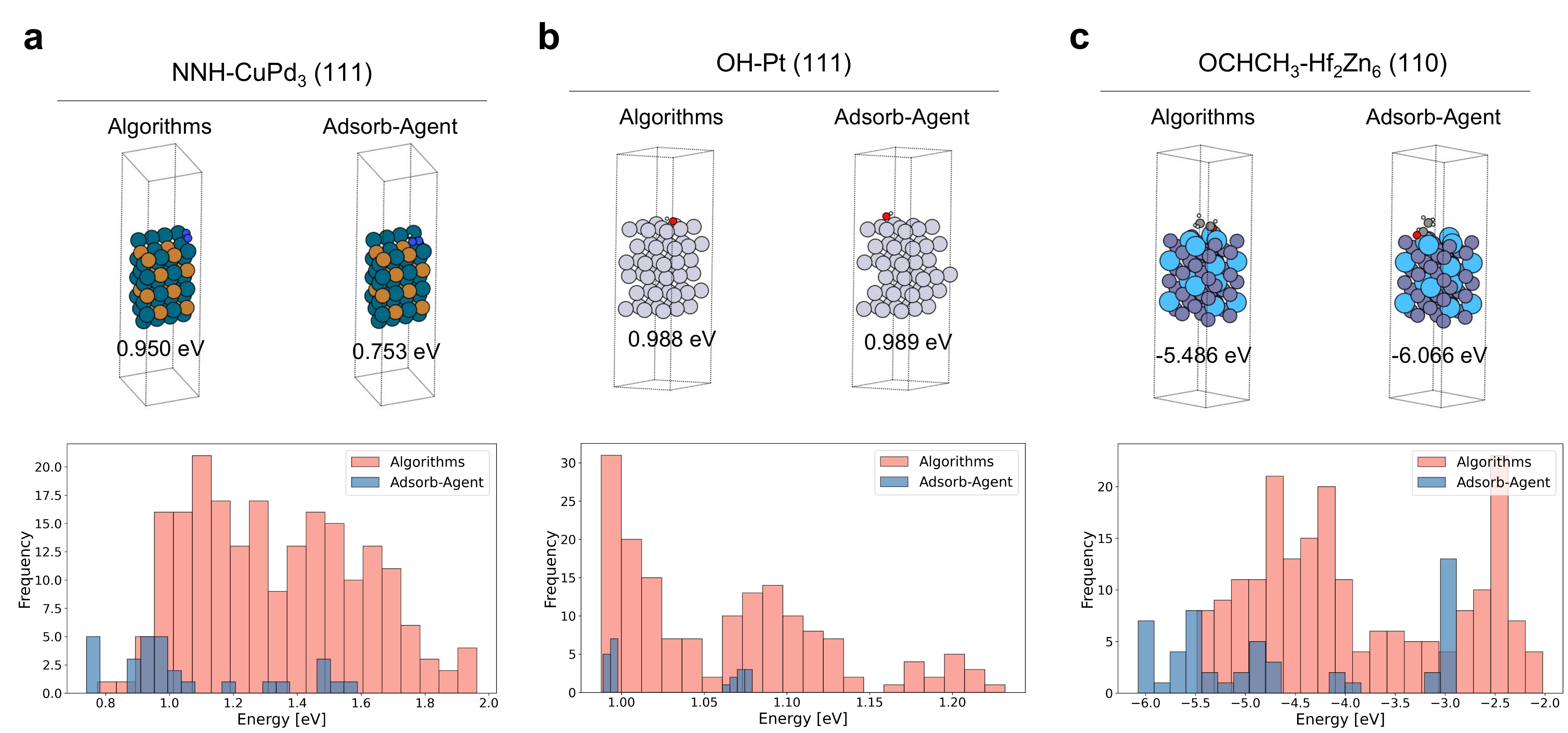} 
\caption{Comparison of Relaxed Adsorption Configurations (upper panels) and Energy Distributions (lower panels). \textbf{a.} \ce{NNH}-\ce{CuPd3} (111); \textbf{b.} \ce{OH}-\ce{Pt} (111); \textbf{c.} \ce{OCHCH3}-\ce{Hf2Zn5} (110). The energy distributions are generated by combining results from all three independent runs.}
\label{fig:visual}
\end{figure*}

As LLMs are inherently non-deterministic, ensuring consistent and reproducible outputs is important. The energy distribution panels in Fig.~\ref{fig:visual} demonstrate that the Adsorb-Agent consistently identifies lower-energy configurations, rather than doing so by chance Notably, for the \ce{NNH}-\ce{CuPd3} (111) and \ce{OCHCH3}-\ce{Hf2Zn6} (110) systems, the frequency of identifying lower-energy configurations is significantly higher compared to the algorithmic approach. This suggests that the Adsorb-Agent effectively targets adsorption configurations closer to the global minimum. For the \ce{OH}-\ce{Pt} (111) system, where the algorithmic approach already exhibits a high frequency of lower-energy identifications, the Adsorb-Agent preserves this trend. These results indicate that the Adsorb-Agent systematically identifies configurations with energies near the global minimum, demonstrating its capability rather than relying on chance.

Furthermore, as shown in Table~\ref{tab:results}, the standard deviations of adsorption energies across multiple implementations of the Adsorb-Agent remain within an acceptable range for most systems. The only exceptions are two systems involving \ce{CH2CH2OH}, where higher deviations are observed. This variability highlights the need for further refinement in handling complex adsorbates while affirming the overall robustness of the Adsorb-Agent.

To quantitatively evaluate consistency across independent trials, we introduce the Consistency Ratio, which measures the proportion of systems that yield consistent solutions across three independent trials. The Consistency Ratio is defined as:

\begin{equation}
\text{Consistency Ratio} [\%] = \frac{\sum_{i=1}^{N} \mathbb{1}(S_{\text{trial}_1, i} = S_{\text{trial}_2, i} = S_{\text{trial}_3, i})}{N} 
\end{equation}

Here, \(N\) represents the total number of systems evaluated, which is set to 20, and \(S_{\text{trial}_j, i}\) denotes the solution obtained for the \(i\)-th system in the \(j\)-th trial.

Consistency is evaluated separately for surface-related and adsorbate-related information. A solution is deemed consistent if the binding atom arrays across the three trials either match exactly or differ by no more than one atom, with the shorter array being a subset of the longer array. This criterion is applied to both surface binding atoms and adsorbate binding atoms using the algorithm detailed in Algorithm~\ref{alg:consistency}. A solution is considered fully consistent only if both surface-related and adsorbate-related information meet these criteria.

\begin{algorithm}
\caption{Consistency Check Algorithm}
\label{alg:consistency}
\begin{algorithmic}
\State \textbf{Initialize:} $\sigma_s \gets \text{True}$ \Comment{$\sigma$: consistency flag}
\ForAll{$a, b \in S$} \Comment{$S$: list of binding atom groups}
    \If{$a \neq b$}
        \If{$|len(a) - len(b)| > 1$}
            \State $\sigma \gets \text{False}$
        \EndIf
        \State $\text{shorter}, \text{longer} \gets \min(a, b), \max(a, b)$ \Comment{Based on length}
        \If{$\text{set}(\text{shorter}) \not\subseteq \text{set}(\text{longer})$}
            \State $\sigma \gets \text{False}$
        \EndIf
    \EndIf
\EndFor
\end{algorithmic}
\end{algorithm}

The Adsorb-Agent demonstrates reasonable reliability, producing consistent solutions for 17 out of 20 systems when the Critic module is applied. Specifically, only one system fails to achieve consistency in adsorbate-related solutions, while two systems fail in surface-related solutions. This highlights the Adsorb-Agent’s strong performance in generating reliable adsorbate-related solutions. Without the Critic module, one additional system fails to achieve consistency in surface-related solutions, indicating the Critic module’s potential role in enhancing solution reliability. Although the limited size of the test set makes it challenging to generalize the effectiveness of the Critic module, these results suggest that the Critic module may help improve consistency by filtering out incoherent solutions within the tested systems.

\subsection{Comparison Across Language Models}

\begin{figure*}[!htb] 
\centering
\includegraphics[width=0.99\textwidth]{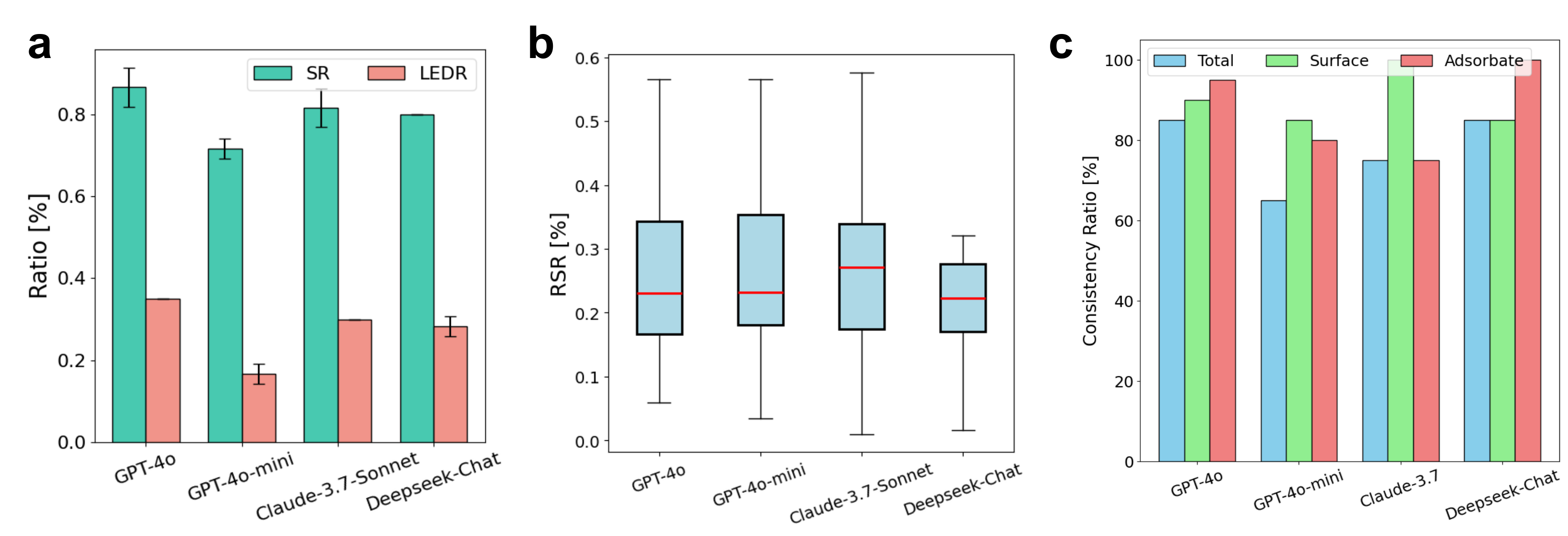} 
\caption{Comparison of Adsorb-Agent performance using different language models. \textbf{a.} Success Rate (SR) and Lower Energy Discovery Rate (LEDR) across models; \textbf{b.} Box plots of Reduced Search Space Rate (RSR) for each language model over three iterations, where the red line indicates the median, the box represents the interquartile range (IQR), and the whiskers extend to the minimum and maximum values within 1.5 times the IQR; \textbf{c.} Consistency ratios for total (surface and adsorbate categories) across independent runs.}
\label{fig:compare_lm}
\end{figure*}

We evaluated the performance of various LLMs, including GPT-4o, GPT-4o-mini, Claude-3.7-Sonnet, and Deepseek-Chat, serving as the ``brain'' of Adsorb-Agent. Among them, GPT-4o demonstrated the strongest overall performance for both SR and LEDR, as shown in Fig.~\ref{fig:compare_lm}a, although its improvements over Claude-3.7-Sonnet and Deepseek-Chat were marginal. In contrast, GPT-4o-mini exhibited the weakest performance, with an average SR of 77\% and an LEDR of 16.7\%. Deepseek-Chat showed the most aggressive reduction of the configuration space, achieving the lowest average RSR of 20.8\%, while the other models yielded average RSR values in the range of 26--28\%, as illustrated in Fig.~\ref{fig:compare_lm}b. Here, the average RSR refers to the mean value calculated across 20 samples over three iterations. Both GPT-4o and Deepseek-Chat attained a high consistency ratio of 85\%, whereas GPT-4o-mini displayed a substantially lower consistency ratio of 65\%, as shown in Fig.~\ref{fig:compare_lm}c. In summary, GPT-4o consistently delivered the best overall performance, while GPT-4o-mini underperformed relative to the other models. Notably, despite being a free and open-source model, Deepseek-Chat achieved performance comparable to the proprietary LLMs.

\section{Conclusion}
We introduced Adsorb-Agent, an LLM-powered agent designed to efficiently explore adsorption configuration spaces and accurately identify adsorption energies. Adsorb-Agent streamlines the computational process by significantly reducing the number of initial configurations required while improving the efficiency and accuracy of minimum adsorption energy predictions, thereby mitigating the computational cost of extensive DFT simulations. By leveraging its built-in understanding of chemical bonding, surface chemistry, and emergent reasoning capabilities, Adsorb-Agent autonomously proposes plausible adsorption configurations tailored to specific systems. Adsorb-Agent demonstrates a strong ability to identify configurations with energies closer to the global minimum, particularly in complex systems such as intermetallic surfaces and large adsorbate molecules. This capability highlights a critical advantage of our approach in addressing computationally intensive and chemically complex scenarios. Our framework underscores the transformative potential of LLM agents in catalysis research, demonstrating how LLMs trained on vast natural language data can effectively assist in solving domain-specific problems.


\section{Methods}
\subsection{GPT-4o}
The GPT-4o model is an optimized variant of OpenAI’s GPT-4, which builds on the advancements of the Generative Pretrained Transformer (GPT) series~\cite{radford2018improving, openai2023gpt4}. Like its predecessors, GPT-4 is a large-scale language model based on the transformer architecture, utilizing self-attention mechanisms~\cite{attention} to effectively model long-range dependencies in text. The GPT-4o model retains the core advantages of GPT-4, including its powerful transformer architecture that ensures high accuracy in language-based tasks. Additionally, GPT-4o has been optimized for more efficient processing, making it well-suited for tasks that demand reduced computational resources without compromising its ability to understand and generate human-like text. This optimization is especially valuable for multi-step reasoning, problem-solving, and decision-making applications. In this study, we used GPT-4o with its default settings: a temperature of 1.0 and top\_p of 1.0.

\subsection{GPT-4o-mini}

GPT-4o-mini is a smaller and more lightweight variant of OpenAI’s GPT-4o model. While it shares the same underlying transformer architecture and core design principles as GPT-4 and GPT-4o~\cite{radford2018improving, openai2023gpt4}, GPT-4o-mini is specifically optimized for faster inference and lower computational costs. This makes it suitable for resource-constrained environments or applications requiring rapid response times. Despite its smaller size, GPT-4o-mini retains the essential capabilities of its larger counterparts, including the ability to perform multi-step reasoning and language understanding tasks. However, its reduced parameter count may lead to trade-offs in accuracy and reasoning depth compared to the full GPT-4o model. In this study, we used GPT-4o-mini with its default settings: a temperature of 1.0 and top\_p of 1.0.

\subsection{Claude-3.7-Sonnet}

Claude-3.7-Sonnet is a large language model developed by Anthropic as part of the Claude 3 family~\cite{anthropic2024claude}. It is based on a transformer architecture, similar to other state-of-the-art language models, and is designed for tasks such as reasoning, summarization, and dialogue. Claude-3.7-Sonnet is positioned as a mid-sized model in the Claude 3 lineup, balancing performance and speed. While its exact number of parameters is not publicly disclosed, it is intended to offer strong capabilities for multi-step reasoning and code generation, with faster responses compared to larger models like Claude-3.7-Opus. Claude-3.7-Sonnet is trained using Anthropic's reinforcement learning from human feedback (RLHF) framework, with particular emphasis on controllability and safety to reduce harmful or biased outputs. In this study, we used Claude-3.7-Sonnet through the Anthropic API with its default settings: a temperature of 1.0 and a top\_p value of 0.999.

\subsection{Deepseek-Chat}

Deepseek-Chat is an open-source large language model developed by DeepSeek AI, released as part of the DeepSeek LLM series~\cite{lu2024deepseekvl}. It is based on a transformer decoder-only architecture, similar to models like GPT, and is trained on large-scale internet datasets to handle a wide range of natural language tasks, including text generation, reasoning, and dialogue. The version used in this study is DeepSeek-V3-0324 7B, which contains approximately 7 billion parameters. Deepseek-Chat adopts techniques such as grouped-query attention and multi-query attention to improve inference efficiency, along with rotary positional embeddings for better generalization to longer sequences. It is trained with supervised fine-tuning and reinforcement learning from human feedback (RLHF) to enhance its instruction-following capabilities~\cite{deepseekai2025}. In this study, we used Deepseek-Chat via the Hugging Face Transformers library with its default generation settings: a temperature of 1.0 and a top\_p value of 1.0.

\subsection{Adsorbate Placement}
A commonly used adsorbate placement involves the \texttt{heuristic} site enumeration algorithm, which leverages surface symmetry~\cite{pymatgen, catkit}. Using Pymatgen’s \texttt{AdsorbateSiteFinder}, this algorithm identifies the most energetically favorable sites, such as ontop, bridge, or hollow sites. The adsorbate is then placed at the selected sites with a random rotation about the z-axis and minor adjustments along the x- and y-axes, ensuring the binding atom is positioned at the site~\cite{oc20}.

To expand the configuration space beyond the \texttt{heuristic} algorithm, Janice et al. proposed a \texttt{random} algorithm~\cite{adsorbml}. This method uniformly samples surface sites at random. After performing a Delaunay triangulation of the surface atoms, random sites are selected within each triangle. At each randomly selected site, the adsorbate is placed with random rotations about the x-, y-, and z-axes, ensuring alignment of the center of mass with the target site~\cite{oc20}.

In this paper, we introduce a new placement strategy specifically designed to accommodate side-on oriented adsorbates, offering a key differentiation from the above methods. Our approach determines the placement center as the weighted center of the binding atoms and orients the adsorbate to maximize the exposure of its binding atoms to the surface. This method differs from the above placement strategies, which position a single binding atom at the site and apply stochastic rotations. Although our strategy operates similarly to existing methods for adsorbates with a single binding atom, it specifically enables side-on placement for bidentate adsorbates.

\subsection{GNN Relaxation}
To evaluate the ground-state properties of an atomic structure, it is necessary to optimize its geometry by minimizing the system’s energy. This optimization is typically achieved by calculating interatomic forces and adjusting atomic positions accordingly, while minimizing the total energy of the system. Traditionally, this process is performed using quantum chemistry methods, such as DFT. However, GNNs have recently emerged as a fast and cost-effective surrogate for DFT-based relaxation~\cite{oc20, adsorbml}.

In GNN-based relaxation, the atomic system is represented as a graph, where atoms are treated as nodes and interatomic bonds or interactions as edges. The GNN iteratively performs message passing, where neighboring atoms exchange information, allowing the network to capture local chemical environments. Through these learned representations, the GNN predicts both the system’s energy and the atomic forces. The atomic positions are then updated using optimization algorithms based on the predicted forces, and this process is repeated until the system converges to a stable, low-energy configuration.

Several GNN models, such as DimeNet++~\cite{dpp}, EquiformerV2~\cite{equiformerv2}, and GemNet-OC~\cite{gemnet-oc}, have advanced the field by incorporating higher-order equivariant representations and scalable optimization techniques. These architectures enable more accurate predictions of energies and forces, resulting in improved relaxation performance across a wide range of molecular and material systems.

After the GNN relaxation, the energy of the adsorbate-catalyst system can be predicted. Since a single adsorbate-surface system may have multiple possible adsorption configurations, it is necessary to evaluate the energies of all candidate configurations to identify the most stable one. The adsorption energy is determined by selecting the configuration with the lowest energy, offering insights into the energetically favorable configuration. The calculation is defined as:



\subsection{Open Catalyst Demo}
The Open Catalyst demo (OC-demo) is an interactive platform that allows users to explore and optimize binding sites for adsorbates on catalyst surfaces~\cite{oc_demo}. It supports 11,427 catalyst materials and 86 adsorbates, with the catalyst crystal structures sourced from the Materials Project~\cite{mp} and the Open Quantum Materials Database~\cite{oqmd}. The OC-demo helps identify the adsorption energy of the selected adsorbate and catalytic surface by generating multiple initial configurations and evaluating them using state-of-the-art graph neural networks, like GemNet-OC and EquiformerV2. These models predict forces for geometry optimization, offering faster results compared to DFT. With support for over 11,000 catalysts and 86 adsorbates, the OC-demo can explore ~100 million surface-adsorbate combinations and perform ~10 billion structure relaxations\cite{oc_demo}. In this study, the results from the algorithmic approach were obtained using this platform, with EquiformerV2 serving as the model for both structure relaxation and energy prediction.

While the GNNs used in the OC-demo represent the state of the art, they show notable discrepancies compared to DFT. However, DFT itself has inherent limitations, including self-interaction errors, challenges in accurately treating strongly correlated systems, underestimation of band gaps, and difficulties in modeling van der Waals interactions~\cite{dft_limit,Burke_2012}. Moreover, additional discrepancies can arise when computational predictions are compared to experimental data, as these models do not explicitly incorporate key factors such as electrochemical charges, solvent effects, temperature fluctuations, or surface imperfections~\cite{Jan2021, Guerin2021}.


\subsection{Evaluation Metrics}

The Success Ratio (SR), estimates the Adsorb-Agent’s ability to identify adsorption energies comparable to those found by the algorithmic approach. Given the inherent error margins in the energy and force predictions by EquiformerV2, a successful identification is defined as the adsorption energy predicted by the Adsorb-Agent falling within a predefined tolerance ($\epsilon$) of the algorithmic approach. The SR is mathematically expressed as:

\begin{equation}
\text{SR} [\%] = \frac{\sum_{i=1}^{N} \mathbb{1}\left(|E_{\text{agent},i} - E_{\text{algorithm}}| \leq \epsilon\right)}{N} \times 100
\end{equation}

Here, \(N\) is the total number of adsorbate-catalyst systems evaluated, set to 20 in this study. The threshold \(\epsilon\) is defined as 0.1 eV, approximately half the energy prediction error of EquiformerV2~\cite{equiformerv2}. \(E_{\text{algorithm}}\) is computed as the average of three independent trials conducted using the algorithmic approaches. The indicator function \(\mathbb{1}(\cdot)\) evaluates to 1 if the condition inside is satisfied and 0 otherwise.

The Lower Energy Discovery Ratio (LEDR), assesses the Adsorb-Agent’s ability to identify adsorption energies that are lower than those found by the algorithmic approaches. It is defined as:

\begin{equation}
\text{LEDR} [\%] = \frac{\sum_{i=1}^{N} \mathbb{1}\left(E_{\text{agent},i} \leq E_{\text{algorithm}} - \epsilon\right)}{N} \times 100
\end{equation}

The Reduced Search Space Ratio (RSR), quantifies the reduction in the number of initial configurations required by the Adsorb-Agent (\(N_{\text{init, agent}}\)) compared to the algorithmic approaches (\(N_{\text{init, algorithm}}\)). A lower RSR indicates a greater reduction in the search space. It is defined as:

\begin{equation}
\text{RSR} [\%] = \frac{N_{\text{init, agent}}}{N_{\text{init, algorithm}}} \times 100
\end{equation}

\section*{Technology Use Disclosure}
ChatGPT was utilized in preparing the preprint version of this manuscript, specifically for assistance with grammar and typographical corrections. All authors have thoroughly reviewed, verified, and approved the content of the manuscript to ensure its accuracy and integrity.

\section*{Code Availability Statement}
The code supporting the findings of this study is publicly available at the following GitHub repository: \url{https://github.com/hoon-ock/CatalystAIgent}.

\begin{acknowledgement}
The authors gratefully acknowledge support from the H. Robert Sharbaugh Presidential Fellowship. We extend our gratitude to Meta Fundamental AI Research (FAIR) Chemistry team for making the Open Catalyst demo service and the Open Catalyst Project dataset publicly available.
\end{acknowledgement}

\section*{Author declarations}
\subsection*{Conflict of Interest}
The authors have no conflicts to disclose.

\subsection*{Author Contributions}
J.O., R.S.M., T.V., Y.J., and A.B.F. designed the research study. J.O., R.S.M., and T.V., developed the method, wrote the code, and performed the analysis. All authors wrote and approved the manuscript.

\bibliography{reference}

\end{document}


\tableofcontents




\newpage
\section{Auxiliary Prompts}

We incorporated two types of auxiliary prompts into the Adsorb-Agent: a reasoning question list for the Solution Planner module and a knowledge prompt for the Critic module. The reasoning question list consists of questions that a human researcher might ask when determining stable adsorption configurations. This list guides the Solution Planner to generate solutions tailored to the task. To ensure the generalizability of the method and to fairly implement the Adsorb-Agent, we intentionally exclude explicit knowledge related to surface chemistry, chemical bonding, or system-specific characteristics. The knowledge prompt, on the other hand, provides terminology clarifications for the Critic module. Its purpose is to ensure that the Critic module can make accurate and contextually appropriate revisions during the solution refinement process.

\begin{figure*}[!hptb] 
\centering
\includegraphics[width=0.99\textwidth]{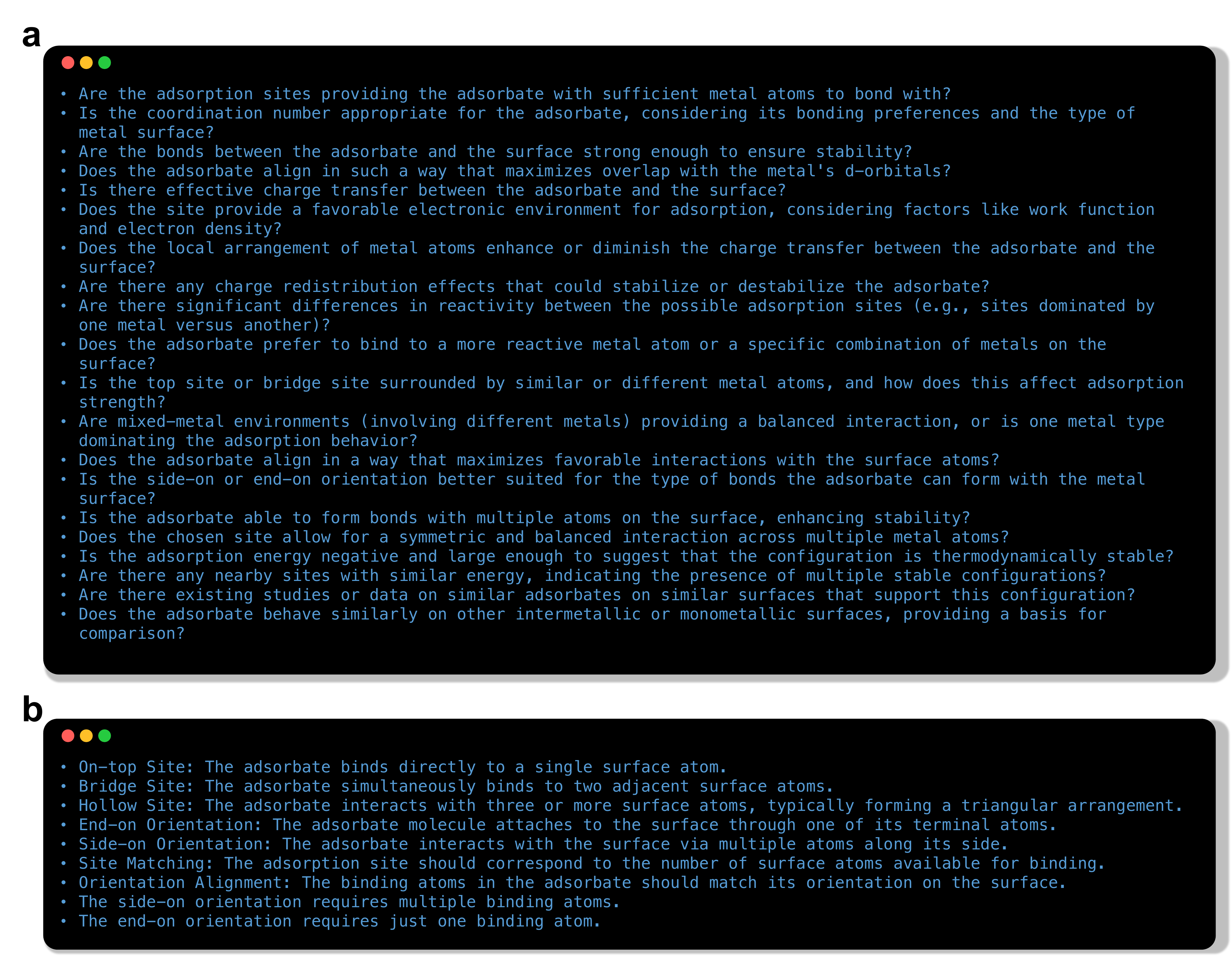} 
\caption{Auxiliary Prompts: \textbf{a.} Reasoning question list, \textbf{b.} Knowledge prompt.}
\label{fig:prompt}
\end{figure*}

\newpage
\section{LLM Modules}

\begin{figure*}[!hptb] 
\centering
\includegraphics[width=0.99\textwidth]{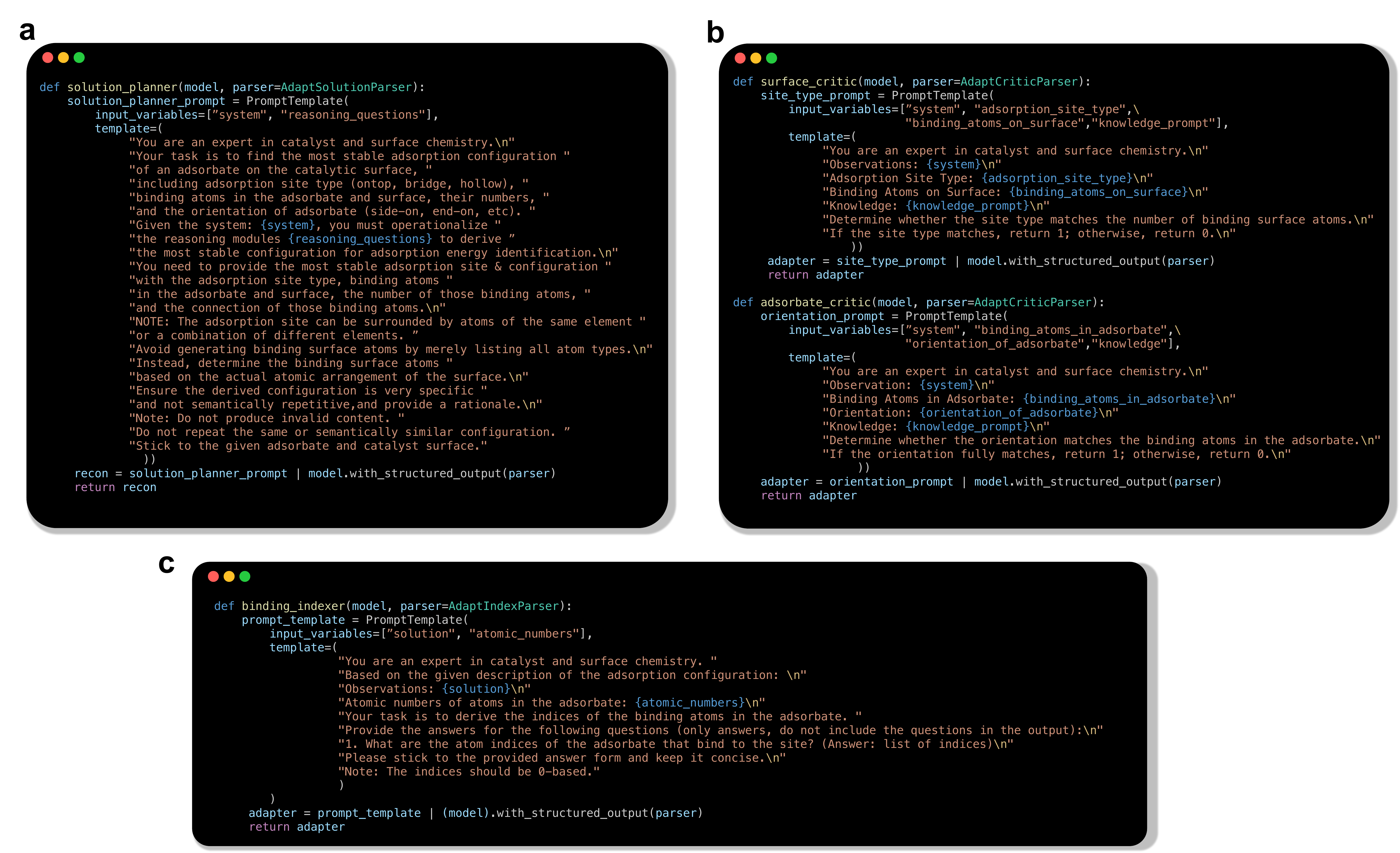} 
\caption{Code Snippets Illustrating the Implementation of the Adsorb-Agent Modules. \textbf{a.} Solution Planner; \textbf{b.} Critic; and \textbf{c.} Binding Indexer.}
\label{fig:code}
\end{figure*}

The Adsorb-Agent consists of three modules powered by GPT-4o, with user-defined templates implemented through prompt engineering. Code snippets illustrating these templates are provided in Figure~\ref{fig:code}. Notably, the Critic module is further divided into two sub-critics: one focused on reviewing surface-related information and the other on adsorbate-related information.

A significant challenge in effectively utilizing an LLM agent is ensuring determinism in the output of a non-deterministic generative LLM. To address this, we incorporated a parser adapter into each module. This adapter modifies the output prompts and extracts the required target information, enabling precise and consistent control over the outputs for our specific applications.

\newpage
\section{Results Across Multiple Runs}

Table~\ref{tab:runs} summarizes the results of individual Adsorb-Agent runs alongside detailed slab metadata. The catalyst bulks are obtained from the Materials Project and are identified by their unique Materials Project ID (mpid). Using the pymatgen package, the selected bulks are sliced along specified Miller indices to generate slab surfaces. The shift parameter determines the position of the cut during slab generation, which defines the surface termination. The top site availability indicates whether adsorption sites are accessible on the topmost layer of the slab.

\begin{table}[!htbp]
\centering
\caption{System metadata and adsorption energies for individual Adsorb-Agent runs. The table provides detailed system metadata, including the adsorbate SMILES, catalyst chemical symbol, Miller index, bulk Materials Project ID (mpid), and key structural parameters (e.g., shift and top site availability). The ``Run" columns display the results from individual Adsorb-Agent runs.}
\label{tab:runs}
\resizebox{\textwidth}{!}{%
\begin{tabular}{lccccccccccccc}
\toprule
\textbf{No.} & \textbf{Adsorbate} & \textbf{Catalyst} & \textbf{Bulk mpid} & \textbf{Shift} & \textbf{Top} & \multicolumn{4}{c}{\textbf{Adsorption Energy [eV]} (\(\downarrow\))} & \multicolumn{4}{c}{\textbf{Number of Initial Sets} (\(\downarrow\))} \\ 
\cmidrule(lr){7-10} \cmidrule(lr){11-14} &  & & &  &  & \textbf{Run 1} & \textbf{Run 2} & \textbf{Run 3} & \textbf{Algorithm} & \textbf{Run 1} & \textbf{Run 2} & \textbf{Run 3} & \textbf{Algorithm} \\ 
\midrule
1   & H   & \ce{Mo3Pd} (111) & mp-1186014 & 0.167 & True & -0.687 & -0.682 & -0.925 & -0.941 $\pm$ 0.002 & 9  & 4  & 7  & 59 \\
2   & NNH & \ce{Mo3Pd} (111) & mp-1186014 & 0.167 & True & -1.370 & -1.041 & -1.383 & -0.903 $\pm$ 0.117 & 14 & 5  & 9  & 51 \\ 
3   & H   & \ce{Pd3Cu} (111) & mp-1184119 & 0.063 & True & -0.379 & -0.384 & -0.377 & -0.398 $\pm$ 0.017 & 18 & 15 & 17 & 98 \\ 
4   & NNH & \ce{Pd3Cu} (111) & mp-1184119 & 0.063 & True &  0.741 &  0.753 &  0.740 &  0.867 $\pm$ 0.072 & 19 & 20 & 13 & 78 \\ 
5   & H   & \ce{Cu3Ag} (111) & mp-1184011 & 0.063 & True &  0.030 & -0.069 & -0.017 & -0.072 $\pm$ 0.002 & 22 & 26 & 16 & 98 \\ 
6   & NNH & \ce{Cu3Ag} (111) & mp-1184011 & 0.063 & True &  1.456 &  1.583 &  1.472 &  1.500 $\pm$ 0.002 & 18 & 13 & 19 & 56 \\ 
7   & H   & \ce{Ru3Mo} (111) & mp-975834  & 0.167 & True & -0.587 & -0.587 & -0.584 & -0.586 $\pm$ 0.050 & 15 & 16 & 20 & 94 \\ 
8   & NNH & \ce{Ru3Mo} (111) & mp-975834  & 0.167 & True & -0.491 & -0.491 & -0.487 & -0.276 $\pm$ 0.003 & 19 & 18 & 19 & 81 \\ 
9   & OH  & \ce{Pt} (111)    & mp-126     & 0.167 & True &  0.989 &  0.991 &  0.991 &  0.990 $\pm$ 0.071 & 7  & 9  & 5  & 54 \\ 
10  & OH  & \ce{Pt} (100)    & mp-126     & 0.250 & True &  0.993 &  0.990 &  0.990 &  0.991 $\pm$ 0.001 & 9  & 16 & 6  & 54 \\ 
11  & OH  & \ce{Pd} (111)    & mp-2       & 0.167 & True &  0.814 &  0.813 &  0.813 &  0.814 $\pm$ 0.001 & 28 & 15 & 17 & 54 \\ 
12  & OH  & \ce{Au} (111)    & mp-81      & 0.167 & True &  1.407 &  1.406 &  1.410 &  1.409 $\pm$ 0.002 & 28 & 16 & 26 & 54 \\ 
13  & OH  & \ce{Ag} (100)    & mp-124     & 0.250 & True &  0.438 &  0.441 &  0.440 &  0.463 $\pm$ 0.009 & 30 & 21 & 20 & 53 \\ 
14  & OH  & \ce{CoPt} (111)  & mp-1225998 & 0.042 & True & -0.218 & -0.188 & -0.220 & -0.166 $\pm$ 0.046 & 43 & 41 & 40 & 120\\ 
15  & \ce{CH2CH2OH} & \ce{Cu6Ga2} (100) & mp-865798 & 0.248 & False & -2.864 & -2.989 & -1.163 & -3.077 $\pm$ 0.062 & 42 & 7  & 37 & 66 \\ 
16  & \ce{CH2CH2OH} & \ce{Au2Hf} (102)  & mp-1018153& 0.028 & False & -1.924 & -3.184 & -3.176 & -3.761 $\pm$ 0.129 & 34 & 27 & 23 & 78 \\ 
17  & \ce{OCHCH3}   & \ce{Rh2Ti2} (111) & mp-2583   & 0.083 & True & -4.570 & -4.562 & -4.552 & -4.275 $\pm$ 0.086 & 23 & 33 & 31 & 62 \\ 
18  & \ce{OCHCH3}   & \ce{Al3Zr} (101) & mp-1065309& 0.375& False & -4.600 & -4.615 & -4.634 & -4.325 $\pm$ 0.052 & 19 & 22 & 25 & 68 \\ 
19  & \ce{OCHCH3}   & \ce{Hf2Zn6} (110) & mp-866108 & 0.120 & True & -6.066 & -5.627 & -6.073 & -5.443 $\pm$ 0.037 & 19 & 15 & 20 & 67 \\ 
20  & \ce{ONN(CH3)2}& \ce{Bi2Ti6} (211) & mp-866201 & 0.000 & True & -3.558 & -3.806 & -3.000 & -2.441 $\pm$ 0.103 & 35 & 36 & 28 & 139\\ 
\bottomrule
\end{tabular}
}
\end{table}

\newpage
\section{System-specific Solutions}

\begin{table}[!htbp]
\centering
\caption{Solutions for the Complete Set of Systems. One trial from three independent runs is shown as an example.}
\label{tab:solutions}
\resizebox{\textwidth}{!}{%
\begin{tabular}{lcccccc}
\toprule
\textbf{No.} & \textbf{Adsorbate} & \textbf{Catalyst} & \textbf{Site Type} & \textbf{Surf. Binding Atoms} & \textbf{Ads. Binding Atoms} & \textbf{Ads. Orientation} \\ 
\midrule
1   & H   & \ce{Mo3Pd} (111) & hollow & Mo, Mo, Pd & H    & end-on \\
2   & NNH & \ce{Mo3Pd} (111) & hollow & Mo, Mo, Pd & N, N & side-on \\ 
3   & H   & \ce{Pd3Cu} (111) & hollow & Cu, Pd, Pd & H    & end-on\\ 
4   & NNH & \ce{Pd3Cu} (111) & hollow & Pd, Pd, Cu & N, N & side-on \\ 
5   & H   & \ce{Cu3Ag} (111) & hollow & Cu, Ag, Cu & H    & end-on \\ 
6   & NNH & \ce{Cu3Ag} (111) & bridge & Cu, Ag     & N, H & end-on \\ 
7   & H   & \ce{Ru3Mo} (111) & hollow & Ru, Mo, Mo & H    & end-on \\ 
8   & NNH & \ce{Ru3Mo} (111) & bridge & Ru, Mo     & N, N & side-on \\ 
9   & OH  & \ce{Pt} (111)    & hollow & Pt, Pt, Pt & O    & end-on\\ 
10  & OH  & \ce{Pt} (100)    & hollow & Pt, Pt, Pt & O    & end-on \\ 
11  & OH  & \ce{Pd} (111)    & hollow & Pd, Pd, Pd & O, H & end-on \\ 
12  & OH  & \ce{Au} (111)    & ontop  & Au         & O    & end-on \\ 
13  & OH  & \ce{Ag} (100)    & ontop  & Ag         & O    & end-on \\ 
14  & OH  & \ce{CoPt} (111)  & bridge & Co, Pt     & O    & end-on \\ 
15  & \ce{CH2CH2OH} & \ce{Cu6Ga2} (100) & bridge   & Cu, Ga & C, O & side-on \\ 
16  & \ce{CH2CH2OH} & \ce{Au2Hf} (102)  & bridge   & Hf, Au & C, O & end-on \\ 
17  & \ce{OCHCH3}   & \ce{Rh2Ti2} (111) & bridge   & Ti, Rh & O, C & side-on \\ 
18  & \ce{OCHCH3}   & \ce{Al3Zr} (101)  & bridge   & Zr, Al & O, C & side-on \\ 
19  & \ce{OCHCH3}   & \ce{Hf2Zn6} (110) & bridge   & Zn, Hf & O, C & side-on \\ 
20  & \ce{ONN(CH3)2}& \ce{Bi2Ti6} (211) & bridge   & Bi, Ti & N, N & side-on \\ 
\bottomrule
\end{tabular}
}
\end{table}
